\pdfoutput=1

\documentclass[11pt]{article}

\usepackage[preprint]{acl}

\usepackage{times}
\usepackage{latexsym}
\usepackage{tabularx}
\usepackage{amsmath}
\allowdisplaybreaks

\usepackage{booktabs}
\usepackage{multirow}
\usepackage[T1]{fontenc}

\usepackage[utf8]{inputenc}

\usepackage{microtype}

\usepackage{inconsolata}

\usepackage{graphicx}

\title{Leveraging Language Models to Detect Greenwashing}

\author{
Margaret Capetz$^1$$^*$ \hspace{2em} Christina Chance$^1$$^*$ \hspace{2em} Rebecca Pattichis$^1$$^*$ \hspace{2em} \\ \textbf{Avalon Vinella}$^1$$^*$ \hspace{2em} \textbf{Reshmi Ghosh}$^2$ \hspace{2em}\textbf{Kai-Wei Chang} $^1$\\
$^1$ University of California, Los Angeles \\
$^2$ Microsoft Research \\
$^*$ Equal Author Contribution \\
\texttt{\{mcapetz17, cchance, pattichi, avinella, kwchang\}@g.ucla.edu}, \\\texttt{reshmighosh@microsoft.com }
}

\begin{document}
\maketitle
\begin{abstract}
In recent years, climate change repercussions have increasingly captured public interest. Consequently, corporations are emphasizing their environmental efforts in sustainability reports to bolster their public image. Yet, the absence of stringent regulations in  review of such reports allows potential greenwashing. In this study, we introduce a novel preliminary methodology to train a language model on generated labels for greenwashing risk. Our primary contributions encompass:  developing a preliminary mathematical formulation to quantify greenwashing risk, a fine-tuned ClimateBERT model for this problem, and a comparative analysis of results. On a test set comprising of sustainability reports, our best model achieved an average accuracy score of 86.34\% and F1 score of 0.67, demonstrating that our proof-of-concept methodology shows a promising direction of exploration for this task.
\end{abstract}

\section{Introduction}
As climate change has moved further to the forefront of social and political discussions, corporations have an increased stake in presenting themselves as environmentally conscious to investors, government agencies, and consumers. Greenwashing is broadly defined as when an entity creates an overly positive corporate image through selective disclosure \citep{bowen2014greenwashing}. In other words, authors of corporate impact reports may present misleading information about their sustainability efforts in order to maintain public favor and profit, rather than being transparent about their potentially negative environmental impact. Natural Language Processing (NLP) can aid with greenwashing detection in the dense language of corporate sustainability reports, and this potential has been recently explored in previous studies. However, a major obstacle is the relative subjectivity of a definition for greenwashing, which is difficult to quantify. In this work, we present methodology for generating ground truth labels for greenwashing risk based on climate-specific attributes. We validate our results with gathered metrics of reported corporate carbon emissions from our collaborators.

\section{Related Work}
We build our work off of previous studies and models that have addressed climate-related issues in an NLP context and defined greenwashing.

\noindent
\textbf{NLP for Climate Work.} 
Many works utilize NLP methods to assess environmental risk and impact. For example, \citet{app12115614} and \citet{polignano-etal-2022-nlp} employ sentence similarity, sentiment classification, and text extraction to evaluate corporate sustainability reports. While \citet{10.1007/978-981-99-1252-0_32} created a search service for greenwashing pattern detection, our work goes beyond pattern detection by developing a novel quantitative definition of greenwashing risk and fine-tuning a specialized language model (ClimateBERT) for this task. Additionally, we validate our results against real-world corporate emissions data.

\noindent
\textbf{ClimateBERT.} 
ClimateBERT is a BERT architecture-based model, more specifically based on Robustly Optimized BERT Pretraining Approach (RoBERTa), pre-trained on a corpus of climate-related texts, which has improved performance on various climate-related downstream tasks \citep{wkbl2022climatebert}. ClimateBERT has been fine-tuned for climate-related tasks such as text detection, sentiment classification, etc \citep{bingler2023cheaptalk}. We build off of this model in our work. 

\noindent
\textbf{Definition of Greenwashing.} 
\citet{doi:10.1525/cmr.2011.54.1.64} define greenwashing as a combination of a firm's positive communication about environmental performance despite true poor environmental performance. Additionally, \citet{ctx14523833220006533} define corporate greenwashing as "the selective disclosure of positive information about a company's environmental or social performance while withholding negative information on these dimensions." This suggests that companies tend to view sustainability information as optional to disclose based on whether or not it will positively impact them. 


\section{Methodology}\label{sec:methods}

\subsection{Data}
\textbf{ClimateBERT Data.}
The ClimateBERT model was pre-trained on 2 million samples of news articles, research abstracts, and corporate climate reports (2015 to 2020) \citep{wkbl2022climatebert}. The public ClimateBERT dataset includes paragraphs sourced from firm's financial disclosures \citep{bingler2023cheaptalk}. We use this data for fine-tuning ClimateBERT for greenwashing risk classification by extracting text from the corpus that was identified as climate-related text from all dataset splits. Then, we re-split into training and validation sets of 80/20 percent (1,056/264 examples), respectively. 

\noindent \textbf{Open for Good Data.} Open for Good is a project at the Center for Impact at the University of California, Los Angeles Anderson School of Management tackling transparent reporting of corporate sustainability \citep{open_for_good}. We use their carbon emissions environmental impact data, which is classified into Scope 1, Scope 2, and Scope 3 by the US Department of Energy. Scope 1 emissions are company’s direct production of a good, Scope 2 includes indirect emissions for purchased electricity, and Scope 3 includes emissions from the value chain of the product's life-cycle. We use these sustainability reports as our test set. 

\subsection{Generating Ground Truth for Greenwashing Risk}
While greenwashing has consistent implications, defining its qualities can be subjective and context-dependent. We have identified the following characteristics typical of greenwashing language in corporate sustainability reports: 1. Absence of explicit climate-related commitments and actions, 2. Use of non-specific language, 3. overly optimistic sentiment, and 4. lack of evasive or hedging terms. This characterization was developed through a comprehensive literature review and corroborated by domain experts from the UCLA Anderson School of Management, ensuring alignment with both academic research and industry insights. Given the legal significance of sustainability reports, there are often instances of legal hedging, or use of evasive language. For example, “The Group is \textit{not aware} of any noise pollution that could negatively impact the environment.” These cases are not necessarily greenwashing, as they are not trying to positively re-frame some adverse action, but more so legal hedging \citep{hedging2017}. 

\citet{lueg2020} propose a method to model the relationship between a company's market risk and its sustainability performance, including a linear combination method to assess various types of risk based on the company's attributes. We therefore adopt a similar approach of using a linear combination of specific metrics to determine greenwashing risk in sustainability-related text.

To quantify the attributes outlined above, we utilize binary labeling, as it is more aligned with the standard sentiments of "positive" and "negative." We utilized pre-trained ClimateBERT models fine-tuned for climate commitment and specificity tasks. These models output probabilities, which we binarized using a threshold of 0.5. For sentiment, while ClimateBERT has been fine-tuned for climate-specific sentiment, the model employs a ternary labeling scheme. To align with our binary labeling, we instead use the pre-trained SiBERT model, which is trained for sentiment analysis \citep{hartmann2023}. The model's output was binarized, with positive sentiment assigned a value of 1 and negative sentiment a value of 0. We developed a custom function to detect hedging language using a predefined list of hedging phrases, see Appendix \ref{appendix: chatgpt_prompt}. The function returns 1 if any hedging phrase is found in the text, and 0 otherwise.

To establish a mathematical formulation for greenwashing, we sourced 10 representative exemplars from the ClimateBERT dataset. Subject-matter experts from the UCLA Anderson School of Management, with extensive background in corporate sustainability and environmental reporting, rigorously annotated and categorized the exemplars, determining whether each exhibited characteristics indicative of greenwashing risk. While this sample size is limited, our work aims to provide a preliminary approach to quantifying greenwashing risk, serving as a proof of concept for future research. The small sample size was due to the time-intensive nature of human-validation. Future work should focus on expanding the ground truth dataset significantly, potentially through data augmentation techniques or other ideas to overcome the challenge of lack of ground truth data.

We used sklearn's least squares linear regression module to fit a linear model to our data. The input features were the four metrics (sentiment, commitment, specificity, and hedging), and the target variable was the expert-annotated label of greenwashing risk. Additionally, utilizing insights around definitions of greenwashing and the characteristics we identified, we iteratively developed another linear combination for comparison. We also provided a threshold to fit the binary nature of the task. 

\section{Results}

\textbf{Ground Truth Labeling.} The linear combination in which coefficients were found through least squares is presented below. Because our final label is binary, we pass the raw output of this equation through a sigmoid, and set a threshold of 0.67 (comes from data-specific observations). The equation exhibits a residual of 0.71. This relatively high residual suggests a suboptimal fit, which could be attributed to the limited sample size used for the regression. Additionally, potential discrepancies between input metrics and output values may play a role, an aspect we will explore further in the subsequent limitations section.

\vspace{-6pt}
\begin{align}
  y = {} & 0.71 \cdot \text{sentiment} + 0.14 \cdot \text{commitment} \nonumber \\
         & - 0.86 \cdot \text{specificity} - 0.71 \cdot \text{hedging}
\end{align}
\vspace{-6pt}

The second formulation, developed from our understanding of greenwashing and metric definitions, is described below. The negative coefficient for sentiment indicates that more positive sentiment increases the risk of greenwashing, which is consistent with the idea that greenwashing often involves overly positive language regarding environmental performance. The positive coefficients for commitment and specificity reflect that companies that are genuinely engaged in sustainability efforts often include explicit commitments and specific language in their reports. The positive coefficient for hedging language may seem counter intuitive, but is based on that fact that companies making genuine environmental claims often use cautious language to avoid over-promising. This formulation aims to capture the essence of greenwashing: overly positive language without specific commitments or details, while also accounting for the nuanced use of hedging in corporate communications.

We received an agreement of 6 out of 10 of the expert human-annotated examples in which non-impact report samples accounted for some incorrect and unclear labels (see Discussion section).

\vspace{-6pt}
\begin{align}
    x &= -\text{sentiment} + \text{commitment} + \text{specificity} \nonumber \\
      &\quad + \text{hedging} \\
    y &= \mathbf{1}_{[x \leq 0]}
\end{align}
\vspace{-6pt}

\noindent \textbf{Fine-Tuning ClimateBERT for Greenwashing Risk.} We fine-tuned the ClimateBERT model (125M parameters) for greenwashing risk detection on data labeled using the first equation, and experimented with freezing all but the classification head vs. fine-tuning all layers. Our hyperparameter search included learning rates (1e-5, 5e-5, 1e-4) and batch sizes (16, 32), with the best performance achieved using a learning rate of 5e-5 and a batch size of 32. We report the mean and standard deviation of validation accuracy and F1 scores across the 10 runs for each configuration. We used the Adam Optimizer, 0 warm up steps, and 70 epochs. Each experiment ran a total of 10 times with different random seeds to report mean and standard deviation (see Appendix \ref{appendix: model_exp} for plots of these results). 

We found that not freezing the RoBERTa layers resulted in higher mean validation accuracy (70.78\% ± 2.03\%) and F1 scores (0.70 ± 0.02) as well as a smaller deviation, compared to the frozen configuration (accuracy: 66.52\% ± 2.39\%, F1: 0.68 ± 0.01). This suggests that the non-frozen model may be slightly more robust than the frozen configuration. It also indicates that our task is specific enough to benefit from training all layers in order to improve the accuracy of ClimateBERT, and that this approach of fine-tuning would be appropriate for determining greenwashing risk.

\begin{table}[h]
\centering
\small
\scalebox{0.93}{
\begin{tabular}{lccc}
\toprule 
\textbf{Model} & 
\bf{\begin{tabular}[c]{@{}c@{}}RoBERTa\\ Layer\end{tabular}} & 
\bf{\begin{tabular}[c]{@{}c@{}}Validation\\ Accuracy\end{tabular}}& 
\bf{\begin{tabular}[c]{@{}c@{}}F1\\ Score\end{tabular}}\\
\midrule

  1 & Frozen & 66.52 $\pm$ 2.39 & 0.68 $\pm$  0.01 \\
  2 & Not frozen & 70.78 $\pm$ 2.03 & 0.70 $\pm$ 0.02 \\

\bottomrule
\end{tabular}
}
\caption{\label{tab:models} Validation results for the frozen vs. not frozen ClimateBERT fine-tuning experiments. This table reports the mean accuracy and F1 scores with their respective standard deviation.}
\vspace{-5pt} 
\end{table}

\noindent \textbf{Out of Distribution (OOD) Evaluation.} We select the run that falls in the mode of accuracy (validation accuracy appears at least twice) and has the highest F1 score within that subset (see Appendix \ref{appendix: model_exp}). Our OOD test set (from the Open for Good Data) come from APA Corporation, Devon Energy, and Diamondback Energy in the oil and gas sector and Autodesk, NVIDIA, and ServiceNow in the tech sector. The sections relevant to environmental impact were manually identified and extracted. To accommodate the maximum input length of the models, these sections were split into paragraph-aligned chunks determined by a character count upper limit before being tokenized. Finally, the report's overall label comes from a majority voting of its intermediate climate-related paragraphs. 

Both models correctly predicted a negative greenwashing risk for each report with significantly higher accuracy than the validation set; however, the F1 variation is much higher (Table \ref{tab:test-accuracies-ground truth}). It is likely that this is due to an imbalance of positive versus negative samples in each report and how they were chunked (see Discussion). 

\begin{table}[h] 
\centering
\footnotesize
\scalebox{0.93}{
\begin{tabular}{lcc|cc}
\toprule 
\multirow{2}{*}{\textbf{Company}} & \multicolumn{2}{c}{\textbf{Model Config 1}} & \multicolumn{2}{c}{\textbf{Model Config 2}} \\
\cmidrule(lr){2-3}\cmidrule(lr){4-5}
& Acc. & F1 & Acc. & F1  \\
\midrule
\textbf{DiamondBack} & 80.70 & 0.62 & 82.46 & 0.67 \\
\textbf{Devon} & 86.67 & 0.65 & 83.33 & 0.63 \\
\textbf{APA} & 86.49 & 0.76 & 85.14 & 0.74 \\
\textbf{Autodesk} & 96.49 & 0.83 & 85.96 & 0.56 \\
\textbf{ServiceNow} & 89.47 & 0.50 & 89.47 & 0.67 \\
\textbf{NVIDIA} & 77.78 & 0.43 & 91.67 & 0.72 \\
\bottomrule
\end{tabular}
}
\caption{\label{tab:test-accuracies-ground truth} Accuracy and F1 scores reported for each company's climate impact statement.}
\end{table}

\section{Discussion}
There are potential improvements in our ground truth generation. For example, the ClimateBERT models are preliminary, and our ground truth for our data is dependent on the accuracy of these models. During human annotation, we noticed the following examples had matching attributes (from ClimateBERT) but were given opposite labels by our domain-experts, see Table \ref{comparison}.

\begin{table}[h]  
\centering
\scalebox{0.5}{
\begin{tabular}{cc}
    \toprule
    \textbf{Text} & \textbf{Label} \\
    \midrule
    \begin{tabular}[c]{@{}l@{}}The Group faces many other risks which, although important and subject to regular review, \\ have been assessed as less significant and are not listed here. These include, for example, \\ natural catastrophe and business interruption risks and certain financial risks....\end{tabular}
      & 1\\
      
     \midrule

     \begin{tabular}[c]{@{}l@{}}
     The cement industry is associated with high CO2 intensity and LafargeHolcim is exposed \\
     to a variety of regulatory frameworks to reduce emissions, some of which may be \\
     under revision. These frameworks can affect the business activities of LafargeHolcim.\\
     In addition, a perception of the sector as a high emitter could impact our reputation, \\
     thus reducing our attractiveness to investors, employees and potential employees.\end{tabular} & 0 \\
     \bottomrule
\end{tabular}
}
\caption{Two examples that had matching attributes but opposite greenwashing risk labels. Attribute values were 1 for sentiment and 0 for the rest.}
\label{comparison}
\vspace{-10pt} 
\end{table}

During annotation, experts noted that while the first example was greenwashing due to obfuscation, the second was not because they are admitting to bad performance. It is these types of nuances that the intermediate attributes may fail to capture. Ultimately, we would prefer to have the ground truth dataset extensively expert-validated. 
\\ \indent Another indicator of greenwashing to consider in our test set is the reported carbon emissions (Scope 1 and 2) for each company relative to its industry's average. If a company has a high risk of greenwashing, their emissions may be higher than average in their domain relative to company size, which we approximate using annual revenue (as in Appendix \ref{appendix:emissions}). Given this data, there might be reason to intuit that APA, whose emissions are incredibly high relative to similar companies, has a risk of using greenwashing language, but neither our ground truth equation nor model reflect this. While it is not definitive evidence, this discrepancy could suggest some error in our formulations. 
 
\section{Conclusion}
We present a novel method of quantitatively defining greenwashing and utilize it to fine-tune ClimateBERT, a BERT model trained on climate-related textual data, to identify risks of greenwashing in corporate sustainability reports. Our best model achieved an average accuracy score of 86.34\% and F1 score of 0.67 on our test set of sustainability reports, which indicates further room for research and improvement in both the ground truth label formulation and training of the language model. 

\section*{Limitations}
To avoid using test data that ClimateBERT had seen in its training, we only used sustainability reports from 2021 and beyond. Because of this, we were limited by the amount of data in the Open For Good data set that met this requirement, skewing our distribution of labels. Table \ref{tab:test-accuracies-ground truth} indicates a potential dataset imbalance, with both models performing well on one label but underperforming on the other. In future work, a balanced dataset can be assembled to provide more valid results. 
\\ \indent Lastly, we are limited to analyzing the sustainability reports paragraph-by-paragraph given the length of the reports. This may impact the final score in both the ground truth generation and model results due to the possible loss of context. Labels may also be impacted depending on text parsing.

\section*{Ethics Statement}
\textbf{Ethical Concerns.} Due to the limitations of our work as well as the inconsistent definition of greenwashing, we utilize the language of potential risk of greenwashing rather than categorizing something as greenwashing itself. With this understanding, our work has no clear ethical concerns as we are not producing data that can be mistreated and have no potential negative applications to the field.

\noindent \textbf{Broader Impact.} Our work provides potential benefits for researchers, policymakers, and corporate stakeholders. For researchers, our proof-of-concept provides a foundation for more sophisticated NLP-based greenwashing detection. This can lead to downstream benefits for policymakers in informing industry environmental regulations and corporate stakeholders like investors, auditors, and corporate sustainability officers in evaluating and improving environmental disclosures. By providing preliminary results for this line of research, we aim to contribute to more transparent, accountable, and impactful corporate sustainability practices and promote environmental responsibility. 


\bibliography{custom}

\appendix

\section{Appendix}
\label{sec:appendix}

\subsection{ChatGPT Prompting Details}
\label{appendix: chatgpt_prompt}
To generate the list of phrases and/or words that we use for tagging a piece of text, we prompted the May 24, 2023 version of ChatGPT. The three turn dialogue prompt used is in Table~\ref{deflection_prompt} as well as the generated hedging phrases. The expanded list of hedging phrases provides more options to capture the nuances of hedging language in a legal setting.This list is used in Python code for various purposes, such as analyzing legal texts or generating automated responses.

Note that this list is a bit broad, as identified by our collaborators. In other words, there are some words that, if present in a text and only that word/phrase is present, it should not necessarily be counted as legal hedging. We discuss improving our greenwashing risk score in our limitations through further hyper-parameter tuning, and this is one area that could benefit from future development.

\subsection{Model Experiments}
\label{appendix: model_exp}

\begin{figure}[h]
    \centering
    \begin{minipage}{0.45\textwidth}
        \centering
        \includegraphics[width=\textwidth]{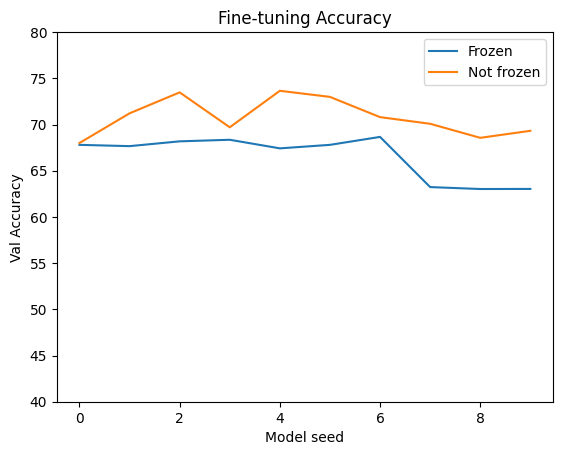}
    \end{minipage}
    \begin{minipage}{0.45\textwidth}
        \centering
        \includegraphics[width=\textwidth]{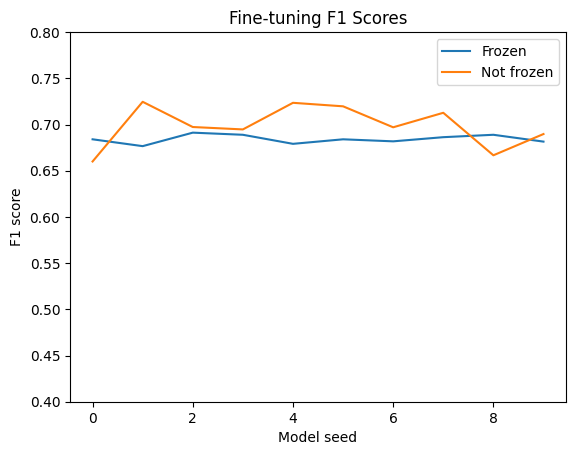}
    \end{minipage}
    \caption{Accuracy and F1 scores of the frozen and non-frozen ClimateBERT models during fine-tuning.}
    \label{fig:evalplot}
\end{figure}
Figure \ref{fig:evalplot} reports the accuracy and F1 scores for each of ten experiments done for frozen and non-frozen training on ClimateBERT. Frozen model 6 and non-frozen model 4 were used to predict the test set.

\subsection{Corporate GHG Emissions and Revenue}
\label{appendix:emissions}
Table \ref{tab:emission-metrics} reports each company's Scope 1 and Scope 2 Carbon Dioxide Equivalents (CO2e) emissions relative to the average emissions of their domain (oil or tech) for that year, as well as their revenue in that same year \citep{open_for_good, wsj_marketdata}.











\begin{table*}[h]
\centering
\footnotesize
\begin{tabular}{l|cccc|c}
\toprule
\multirow{ 2}{*}{\textbf{Company}} & \multirow{2}{*}{\textbf{Scope 1 CO2e}} & \multicolumn{3}{c|}{\textbf{Scope 2 CO2e (relative to average)}} & \multirow{2}{*}{\textbf{Annual Revenue}} \\
\cmidrule{3-5}
& & Market & Location & Uncategorized & \\
\midrule
\textbf{DiamondBack} & -20 & - & +0 & - & 2.8 \\
\textbf{Devon} & +6.9 & - & - & -0.36 & 4.8 \\
\textbf{APA} & +31 & - & - & +1.3 & 4.0 \\
\textbf{Autodesk} & -0.34 & -0.30 & -0.47 & - & 3.8 \\
\textbf{ServiceNow} & -0.33 & -0.11 & -0.30 & - & 4.5 \\
\textbf{NVIDIA} & -0.32 & +0.36 & +0.18 & - & 17 \\
\bottomrule
\end{tabular}
\caption{\label{tab:emission-metrics} Each company's reported Scope 1 and Scope 2 emissions measured in hundred thousand metric tonnes, relative to the hundred thousand average emissions of their domain (oil or tech) for that year, as well as their revenue in billions of USD.}
\end{table*}

\begin{table*}[h]

\footnotesize
\begin{tabularx}{1.0\textwidth}{X}
    \toprule
    \textbf{User}  Generate a list of phrases or words that represent deflection languages. Examples include: "not aware", "unsure"\\
    \textbf{ChatGPT} 
    \begin{enumerate}
    \itemsep0em 
        \item I'm not certain.
        \item I can't say for sure.
        \item I'm not entirely convinced.
        \item It's hard to say.
        \item I'm not familiar with that.
        \item I don't have all the information.
        \item ...
    \end{enumerate}
    \textbf{User} Generate a list of phrases or words that represent deflection languages seen in legal language. Examples include: "not aware", "unsure":\\
    \textbf{ChatGPT} 
    \begin{enumerate}
        \itemsep0em 
        \item Alleged
        \item Purported
        \item Supposed
        \item Ostensibly
        \item Arguably
        \item Presumably
        \item Putatively
        \item ...
    \end{enumerate}
    \textbf{User} Can you generate the above list as a Python list of string?\\
    \textbf{ChatGPT} Certainly! Here's the list of deflection phrases in legal language as a Python list of strings:

    \textit{deflectionPhrases} = [
    "Alleged",
    "Purported",
    "Supposed",
    "Ostensibly",
    "Arguably",
    "Presumably",
    "Putatively",
    "Seemingly",
    "Reportedly",
    "Apparent",
    "So-called",
    "Allegedly",
    "Potentially",
    "Possibly",
    "Putative",
    "Suggested",
    "Circumstantially",
    "Put forth",
    "By all accounts",
    "Inferred"]\\

    \textbf{User} Can you append more words to the previous list related to deflection languages for the legal setting?\\
    \textbf{ChatGPT} Certainly! Here's an extended list of deflection phrases in legal language:\\
    \textit{deflectionPhrases} = [
    "Alleged",
    "Purported",
    "Supposed",
    "Ostensibly",
    "Arguably",
    "Presumably",
    "Putatively",
    "Seemingly",
    "Reportedly",
    "Apparent",
    "So-called",
    "Allegedly",
    "Potentially",
    "Possibly",
    "Putative",
    "Suggested",
    "Circumstantially",
    "Put forth",
    "By all accounts",
    "Inferred",
    "Conceivably",
    "Inconclusively",
    "Tentatively",
    "Speculatively",
    "Hypothetically",
    "In all likelihood",
    "Plausibly",
    "Implied",
    "Indirectly",
    "Presumptively",
    "Theoretically",
    "Presumed",
    "Conditionally",
    "Preliminarily",
    "Provisionally",
    "Notoriously",
    "Put forward",
    "Understood",
    "It is said",
    "Subject to interpretation",
    "Perceived",
    "In a sense",
    "To a degree",
    "Assumed",
    "Putatively",
    "Vaguely",
    "Broadly",
    "Loosely",
    "Dubiously",
    "Evidently",
    "Inferred",
    "Outwardly",
    "Reputedly",
    "Tacitly",
    "Unofficially",
    "Putatively",
    "Ostensible",
    "Presumptuous",
    "Unconfirmed"
]\\
    \bottomrule
    \end{tabularx}
 \caption{\label{deflection_prompt} Three turn dialogue to generate the list of phrases and/or words that we use for tagging a piece of text, we prompted the May 24, 2023 version of ChatGPT.}        
\end{table*}

\end{document}